\documentclass{osa-article}

\journal{osajournal}

\articletype{Research Article}
\usepackage{graphicx}
\usepackage{algorithm}
\usepackage{algpseudocode}
\usepackage{amsmath}
\usepackage{graphics}
\usepackage{epsfig}
\usepackage{booktabs} 
\begin{document}

\title{A new multilayer optical film optimal method based on deep q-learning}

\author{A.Q. Jiang,\authormark{1,*} Osamu Yoshie,\authormark{1} and L.Y. Chen\authormark{2}}

\address{\authormark{1}Graduate school of IPS, Waseda University, Fukuoka 8080135, Japan\\
\authormark{2}Department of Optical Science and Engineering, Fudan University, Shanghai, 200433, People’s Republic of China}

\email{\authormark{*}anqingjiang0524@akane.waseda.jp} 



\begin{abstract}
Multi-layer optical film has been found to afford important applications in optical communication, optical absorbers, optical filters, etc. 
Different algorithms of multi-layer optical film design has been developed, as simplex method, colony algorithm, genetic algorithm. 
These algorithms rapidly promote the design and manufacture of multi-layer films. 
 However, traditional numerical algorithms often converge to local optimum. 
 This means that these algorithms can not give a global optimal solution to the material researchers.
 In recent years, due to the rapid development of artificial intelligence, to optimize optical film structure using AI algorithm has become possible.
 In this paper, we will introduce a new optical film design algorithm based on the deep Q learning. 
 This model can converge the global optimum of the optical thin film structure, this will greatly improve the design efficiency of multi-layer films.
\end{abstract}

\section{Introduction}
After the 1970s, with the development and application of computer technology and various numerical optimization techniques. 
Many traditional numerical optimization algorithms, such as linear programming, simplex method, least square method, damped least squares 
method were applied
 to the design of optical thin film. In which the least square method was the most successful application. The shortages of this several method are,
 this methods are focus on find the local optimum value of the film coating. The optical film researchers
  also developed a variety of global optimal methods that can be optimized in the film structure.
Tikhonravov, Alexander V., and Michael K developed a optical coating design software system based on needle optimization technique. \cite{tikhonravov1994development} 
Sullivan, Brian T., and J. A. Dobrowolski. implemented the needle optimization method, and improve design system performance, 
The improved needle algorithm can not only define very complex evaluation functions but also can be used in many forms to calculate and design, 
It made the design of optical thin film becomes more flexible.\cite{sullivan1996implementation} C.P. Chang and H.Y. Lee apply the GSAM(generalized simulated-annealing method)
thin-film system design and find that there is no local minimal trapping problem in 1990.\cite{chang1990optimization}. Martin, S applied
 genetic algorithms to the design of three very different optical filters.\cite{martin1995synthesis} In recent decade, this field researcher
 develop the optical coating optimization system based on several model as, GA, ant colony algorithm, particle swarm optimization and so on.\cite{paszkowicz2013genetic}\cite{tang2006coating}\cite{guo2014design}

In past several years, machine learning was widely used in many research field, as  computer vision, automatic robot, medical, finance, etc. It make a big change for our society.
\cite{michalski2013machine} Thanks to the growth of computing power, many deep learning methods have been developed by scientists and researchers to solve problems in various fields.
The deep reinforcement learning(DRL) is a successful method, which refers to goal-oriented algorithms. It means DRL could learn how to attain a complex objective (goal) or maximize along a particular dimension over many steps.\cite{mnih2015human}
The famous application of DRL is Alpha-Go.\cite{Silver2017Mastering}
This research use a deep q network, one kind of a deep reinforcement learning method, to optimal the optical coating. We optimal 2 different optical coating(anti-reflection film
and solar selective absorption film) use our proposed method to measure the effectiveness of this method in this paper. 

\section{Deep Q learning}
Deep Q learning combines reinforcement learning with a class of artificial neural network known as deep neural networks.
There are two basic elements of reinforcement learning: Agent and Environment. Agents act to influence the environment, 
and then the environment return a feedback to the agent. According this feedback the to decide the next action.
Fig.1 is a schematic diagram of the process of  reinforcement learning

\begin{figure}[htbp]
\centering\includegraphics[width=10cm]{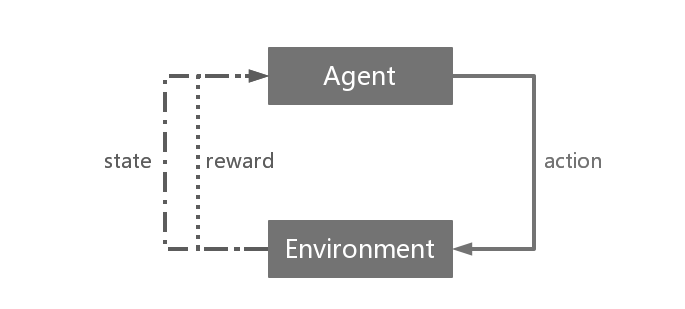}
\caption{Reinforcement Learning process}
\label{fig_brain_env}
\end{figure}
\subsection{Markov decision process}

It is called Deep Q learning, because the author used a deep neural network to replace a Q-table. Q-table the kernel method in Q
 learning\cite{Watkins1989Learning}. In short, deep Q learning develops control patterns by providing feedback on a model’s selected
actions, which makes the model to select better actions in the next step. 

Q-learning is a solution to a Markov decision process. A Markov decision process is defined by an agent, performing in an
 environment by means of actions, as showed in Fig.2. The environment can be in different states. Markov decision process is a 4-tuple
 $(S, A, P_{\alpha}, R_{\alpha})$. $S$ is a finite set of states, $A$ is a finite set of actions, $P_{\alpha}$ is a transition
 probabilities, that action $\alpha$ i state $s$ at time $t$ change to state $s_{'}$ at time $t+1$. $R$ is the reward received,
  when state $s$ transfer to  state $s_{'}$ under the influence of action $\alpha$.\cite{Bellman1957A}\cite{Altman1995Constrained}
 
\begin{figure}[htbp]
  \centering\includegraphics[width=7cm]{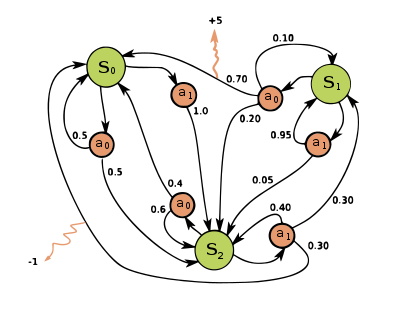}
  \caption{Example of a simple MDP with three states(green circles) and two actions(orange circles), with two rewards(orange arrows)}
  \label{MDPs}
\end{figure}

  The set of states and actions, together with rules for transitioning from one state to another and for getting rewards, make up a Markov decision process. One episode
  of this process forms a finite sequence of states, actions and rewards:
  \begin{equation}
	s_{0}, a_{0}, r_{1}, s_{1}, a_{1}, r_{2}, s_{2}, …, s_{n-1}, a_{n-1}, r_{n}, s_{n}
  \end{equation}
 Here $s_{i}$ is the state, $a_{i}$ is the action and $r_{i+1}$ is the reward after performing the action. The episode ends with terminal state $s_{n}$
 
\subsection{Discounted Future Reward}
Considering the long-term situation, not only immediate reward need to be token into account, but also the future awards should be considered.
The total reward for one episode in a Markov decision process could be easily calculate as:
\begin{equation}
	R=r_{1}+r_{2}+r_{3}+…+r_{n}
\end{equation}
The total future reward from time $t$ could be expressed as:
\begin{equation}
	R_{t}=r_{t}+r_{t+1}+r_{t+2}+…+r_{n}
\end{equation}
Because our environment is stochastic, if every step of action will make a same reward. For long-term situation, it may
 diverge. For that reason it is common to use discounted future reward instead:
 \begin{equation}
	R_{t}=r_{t}+\gamma (r_{t+1}+\gamma (r_{t+2}+…))=r_t+\gamma R_{t+1}
 \end{equation}

Here $\gamma$ is the discount factor between 0 and 1. The more into the future the reward is, the less we take it into consideration. It is easy to see, that discounted 
future reward at time step t can be expressed in terms of the same thing at time step $t+1$.

\subsection{Q-learning}
Q learning is a model, which approximates the maximum expected return for performing an action at a given state using an 
action-value Q function. It represents the discounted future reward when we perform action $a$ in
state $s$.The Q function is defined as,
\begin{equation}
  Q(s,\alpha) = r + \gamma max(Q(s^{'}, \alpha^{'}))
\end{equation}
where $r$ is the reward that was providing feedback from environment, $s$ is the state, $\alpha$ is the action, $\gamma$ is learning rate, and $Q(s^{'}, \alpha{'})$
 comes from predicting the Q function for the next state using our current model.The main idea in Q-learning is that we can iteratively approximate the Q-function 
 using the Bellman equation. In the simplest case the Q-function is implemented as a table, with states as rows and actions as columns. This talbe is called q-table.

\subsection{Deep Q-network}
The state of the real situation may be infinite, so that q-table will be infinite. It means the even a supercomputer also can't calculate
the solution of this q-table, when the data or environment is complex. As we know, neural networks are good in optimal highly structured data.
We could represent our Q-function with a neural network, that takes the state and action as input and outputs the corresponding Q-value.
Fig.3 shows the transformation of Q-learning to deep Q-learning.
\begin{figure}[!h]
  \centering
  \includegraphics[width=2.5in]{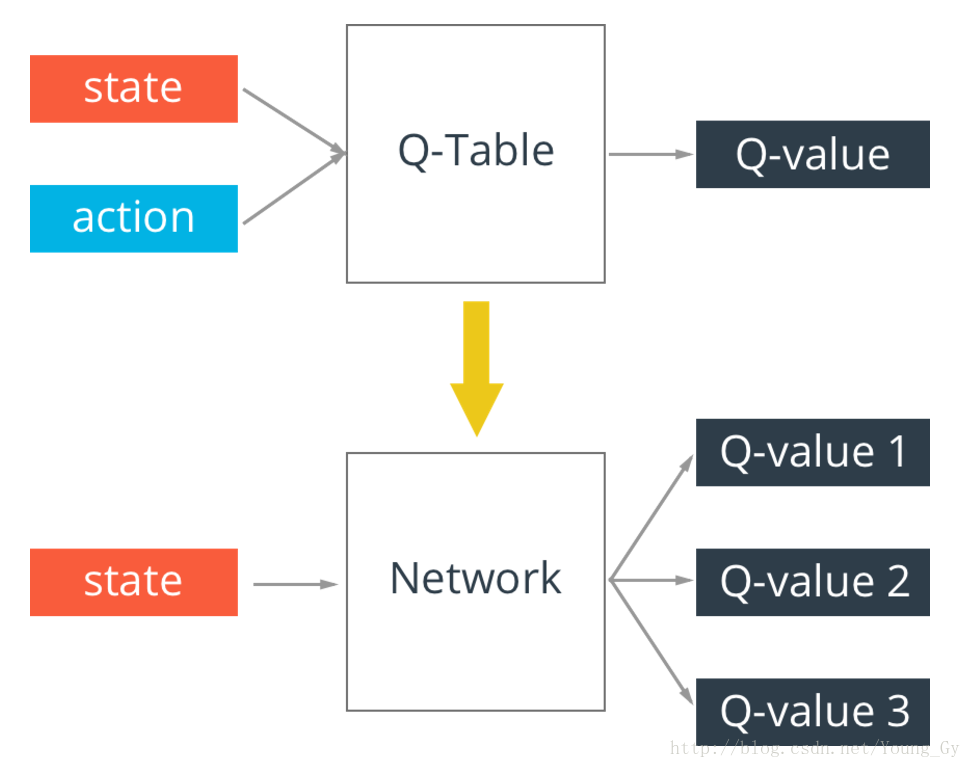}
  \DeclareGraphicsExtensions.
  \caption{Top: The Q-learning model. Bottom: the deep Q-network, used in DeepMind paper.}
  \label{DQN}
\end{figure}

\subsection{Loss functions}
In most parts of training process, Q-values can be any real values or any real integer. It determines by the model you want to
optimize. So the neural network of deep Q-learning can be optimized with a simple squared error loss.
\begin{equation}
	Loss=\frac{1}{2}[r + \gamma max_{a'}Q(s',a')-Q(s,a)]^2
\end{equation}
$r + \gamma max_{a'}Q(s',a')$ represents the target value. $Q(s,a)$ represents the prediction value.
\subsection{Experience reply}
Because the we use neural network to approximate the Q-table in Deep Q-learning, so it is not stable in many situation.
It will cause the model can not converge or converge to the local optimal value. There are several way to solve this problem.
The experience replay is one of the most useful way. When training the neural network, We use random mini-batch from the
memory to instead of the recent transition. This breaks the similarity of the training samples, which make our network
into a local minimum.

\subsection{Exploration exploitation}
As the temperature in  annealing algorithm, a higher freedom in selecting actions at the beginning is necessary. The model
will random select action with probability $\epsilon$.The $\epsilon$ will between 0.1 to 1, and it will decrease with the
 training process. This technology allows the model to converge faster and avoid convergence in the local optimal.

\subsection{Update algorithm}
Combining the above methods, the Q-network can be optimized as follow algorithm:

\begin{algorithm}[htb]
  \label{alg:DQN}
\caption{Update algorithm of deep Q learning}   
\begin{algorithmic}[1] 
\State initial the memory $D$
\State initial the $Q$ network with random weights $w$
\State select a random action $a_{t}$ with probability $\epsilon$, otherwise select $a_{t} = argmax_{a}Q(s, a)$ 
\State Execute action $a_{t}$ in the system environment, observe reward $r_{t+1}$ and new state $s_{t+1}$
\State Store transition $<s_{t}, a_{t}, r_{t+1}, s_{t+1}>$ in memory $D$
\State Sample random mini-batch from $D:<s_{j}, a_{j}, r_{j},s_{j}^{'}>$
\State Set $Q_{j} = r_{j} + \gamma max_{a}Q(s_{j}^{'}, a{j}^{'})$
\State Gradient descent step with a simple squared error loss.

\For {episode = 1, $M$}
\For {$T, t$}
\EndFor
\EndFor
\end{algorithmic}
\end{algorithm}

\section{Application to the Multi-layer Systems}
To apply deep Q-network on multi-layer film optimize, we define the state, action, and reward of multi-layer
system.
\subsection{States}
 There are several different layer in one multi-layer optical film, each layer have different
  thickness and different thickness. We define the layer thickness of multi-layer optical film as a state 
in the optimal system, as:
\begin{equation}
s=Th_{1}, Th_{2},..,Th_{n}
\end{equation}
For example, in a 4 layer solar absorber film, which structure is
$SiO2(90nm)/Cr(10nm)/SiO2(80\\nm)/Al(\geq200nm)$.We will define the state as an array\emph{[90,10,80]} .
This array value is according the thickness of each layer.The $Al$ layer is a layer which keep
 transmission equal to zero in solar absorber film. To reduce the calculation, the $Al$ layer will 
 be ignored. 

\subsection{Actions}
Inspiration from the needle algorithm, the actions defined as increasing or reducing the thickness of 
one layer in the film coating. The minimum thickness change is determined by the design precision of 
the film coating. For example, in a $n$ layers film coating with a design accuracy of 0.01 nm, the action will be
defined as:
\begin{equation}
  A=\{a_{1}, a_{2},..,a_{n}\}
\end{equation}

$a_{n}$ represents the action on each layers. In this situation $a_{n}$ contains 6 values, each value is an
action on the layer $n$ in this film coating.
\begin{equation}
  a_{n}=\{x|\pm x=10^{-K}, 0\geq K\geq N, K\in Z^{+}\}
\end{equation}

$N$ represents the precision level of this film coating.
In this case,the $N$ is equal $3$, that $a_{n}$ will equal ${\pm 1,\pm 0.1,\pm 0.01}$. We choose several
different thickness change is aim to speed up the convergence speed while guaranteeing the accuracy. It's 
very easy to understand, an action $a_{n} = 1$ will save a hundredfold of the time compared with $a_{n}=0.01$

\subsection{Reward}
Before define the reward, the aim function of film coating optimal should be defined at first.
There are three parameters, transmission $T$,  reflection $R$, absorption $A$, are focused in the design of optical film coating design.
The reward in our film coating optimal system is defined as 3 value, as:
\begin{equation}
  Aim = w_{T}T + w_{R}R + w_{A}A
\end{equation}

$w_{T},w_{R}, w_{A}$ represent the weights of three parameters. This three weights determined by type of 
optical thin film. For example, the weights will defined as $w_{T}=0,w_{R}=-1,w_{A}=1$ during designing the
solar selective absorber. Because influence of transmission can be ignored, and film need higher absorption
 and smaller reflection.
\begin{equation}
  r_{t} = \begin{cases}
    0, \text{not find a better value in past 50 steps} \\
    -1, \text{aim optical parameters is lower threshold value} \\
    Aim_{best} - Aim_{t}, {other situation}
  \end{cases}
\end{equation}

We define a stop search mechanism, when the system not search a better result in several steps. 
It has the chance to reduce the convergence probability of the model, but it can get the best 
value after all. Compared with other application of deep Q-learning, our system pays more attention
 to finding the best solution. 

\subsection{Neural network structure}
Considering the number of actions, we use a two layers full connect neural network in our deep Q-network.
Each layer has 80 units.We will random initialize this network. The learning rate will fade with the 
training of the network. 

\section{Experiments}

In this part, we will show the application performance in 2 different optical
coatings(solar selective absorber and high reflection film).

\subsection{Solar selective absorber}
First we apply deep Q-learning model on $300nm$-$1500nm$ solar selective absorber. 
4-layers, 6-layers film structure is very famous structure in
solar selective absorber. The materials of film coating use $Ti$ and $SiO_2$. The
substructure will use $200nm$ $Cu$. The thickness is initialized with $50nm$ for 
each layer. 
\begin{table}[H]
  \centering
  \caption{Solar selective absrober coating after optimized}
\begin{tabular}{ccccc}
  \toprule 
  layer&material& 4-layers& 6-layers& 8-layers\\
  \midrule  
  0 &Air   & -    & -& -\\
  1 & $SiO_{2}$ & 132.4& 126.0& 63.19 \\
  2 & $Ti$   & 13.74  & 6.46& 3.47\\
  3 & $SiO_{2}$ & 77.5& 73.37& 71.46\\
  4 & $Ti$   & - & 12.98& 6.19\\
  5 & $SiO_{2}$ & - & 54.56& 65.84\\
  7 & Ti   & - & - & 12.45\\
  8 & $SiO_{2}$  & - & - & 52.4\\
  Sub & $Cu$  & 200  & 200 & 200\\
  \midrule  
  Absorption & & 87.4\%& 90.15\%& 91.18\%\\
  \bottomrule
\end{tabular}
\end{table}

\begin{figure}[H]
  \centering\includegraphics[width=8cm]{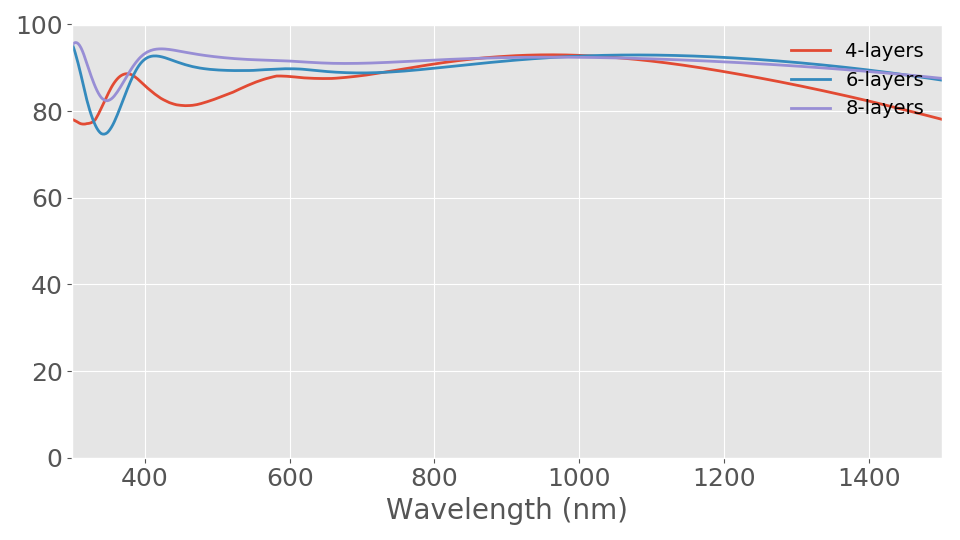}
  \caption{The optimal result of DQN algorithm.}
\end{figure}

The Table 1 and Fig.4 shows the optimal result of film coating. The initial thickness of
each layer could be random. But we do not hope you initialize it very large, it will
make you cost many time to get the final good optimal result. In experiments, if the 
initial film coating is very close to the optimal result it will find the best result
very fast. By contrast, it costs much time. But in the end, it will find best optimal
result almost irrelevant to the initial value.

\subsection{Anti-reflection film }
Anti-reflection film is also a very important application in optical film. The anti-reflection
film applies on camera lens, solar absorber and so on. In this part, we test a $ZnS$ and $MgF_{2}$
anti-reflection film on $400nm$ to $700nm$. In this experiments, the precision of design is $1nm$. The 
initial thickness of each layer is $30nm$.So the number of actions is come to equal 18 (9 layers 
time 2 action of each layer). We compared the result with another optimal technique: GA. 

\begin{table}[H]
  \centering
  \caption{Anti-reflection film coating after optimized}
\begin{tabular}{cccc}
  \toprule 
  layer&material&  DQN & GA\\
  \midrule  
  0 &Air         & -   &\\
  1 & $MgF_{2}$  & 96  &90\\
  2 & $ZnS$       & 27  &23\\
  3 & $MgF_{2}$  & 14  &16\\
  4 & $ZnS$        & 70  &71\\
  5 & $MgF_{2}$  & 22  &20\\
  7 & $ZnS$        & 24  &26\\
  8 & $MgF_{2}$  & 43  &40\\
  9 & $ZnS$        & 7   &3\\
  Sub & $SiO_{2}$ & -& -\\
  \midrule  
  Reflection & &4.5\% & 5.9\%\\
  transmission & &94.0\% & 92.7\%\\
  \bottomrule
\end{tabular}
\end{table}

According the experiments, we could know the Deep Q-learning could find better solution compare with 
GA in same situation. The spectral curve shows in Fig.5. The reflection distribution of DQN is more
uniform.

\begin{figure}[H]
  \centering\includegraphics[width=14cm]{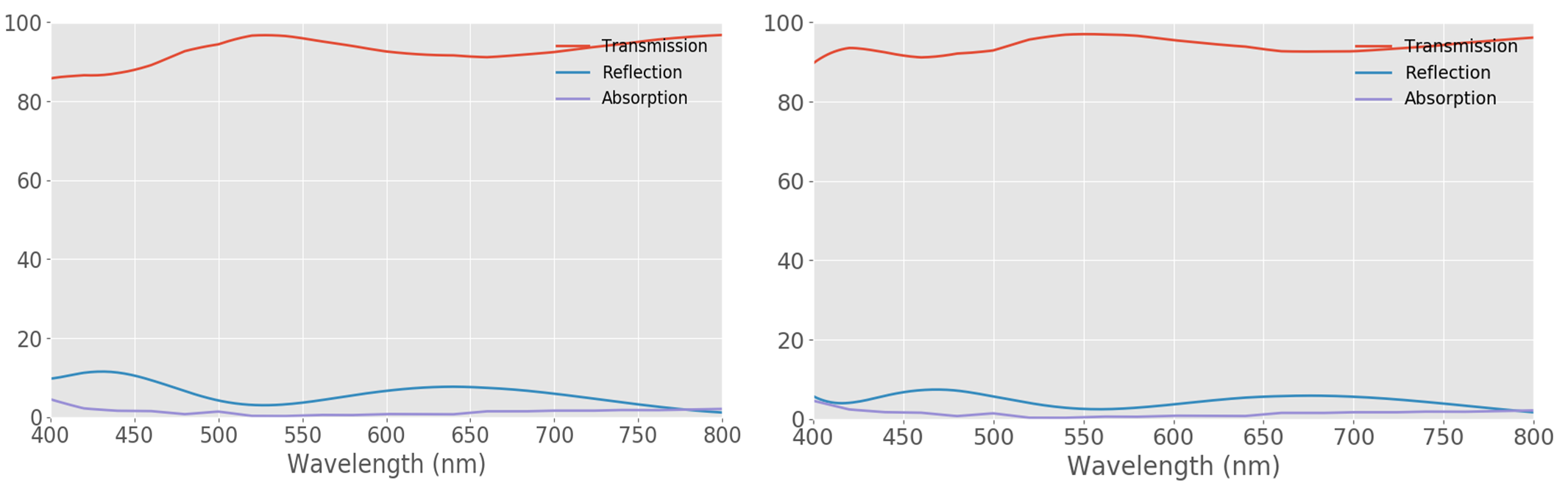}
  \caption{Left: the optimal result of GA algorithm. Right: the optimal result of DQN algorithm.}
\end{figure}

\section{Conclusion}

This paper provides a new method for numerical optimization of optical thin film design.
This new method combines machine learning and optical thin film computing. Compared with
 previous research in the same field, this algorithm has higher search space and better 
 search results. This algorithm can reach the search scope of needle algorithm and the
  local search precision of GA algorithm. 
  
In our experiments, the initial thickness of the optical thin film could be random. 
This means that researchers can select materials and perform parallel operations 
on a large scale according to their requirements. This will greatly improve the efficiency
 of finding the target film.

\section{Acknowledgment}

The authors would like to thank Prof.Chen from FuDan Univ. and Prof.Yoshie from Waseda Univ..
These two professors give me a lot of research advice and provided me with a good research 
environment. The optical film computation part is based on C. Marcus Chuang's open source project
\cite{OpticalModeling}. MorvanZhou give us a great hello wolrd demo of deep 
Q-learning\cite{Morvan}. The optical constant data is search from refractive index.info this 
website built by Mikhail Polyanskiy\cite{RefrectInfo}. In the end, I want to thank all those who have given me
 help in the study of life.

\bibliography{OSA-template}






\end{document}